\documentclass{article}

\PassOptionsToPackage{numbers,sort&compress}{natbib}

\usepackage[preprint]{neurips_2019}

\usepackage[utf8]{inputenc} 
\usepackage[T1]{fontenc}    
\usepackage{hyperref}       
\usepackage{url}            
\usepackage{booktabs}       
\usepackage{amsfonts}       
\usepackage{nicefrac}       
\usepackage{microtype}      
\usepackage{amsmath}
\usepackage{amssymb}
\usepackage{graphicx}
\usepackage[font=it]{caption} 
\usepackage{tabularx}
\usepackage{wrapfig}
\usepackage{adjustbox}      
\usepackage{verbatim}
\usepackage{textcomp}       

\usepackage[dvipsnames]{xcolor} 
\usepackage[normalem]{ulem} 

\newcolumntype{L}[1]{>{\raggedright\let\newline\\\arraybackslash\hspace{0pt}}b{#1}}
\newcolumntype{C}[1]{>{\centering\let\newline\\\arraybackslash\hspace{0pt}}b{#1}}
\newcolumntype{R}[1]{>{\raggedleft\let\newline\\\arraybackslash\hspace{0pt}}b{#1}}

\title{Symmetry-adapted generation of 3d point sets for the targeted discovery of molecules}

\author{
	Niklas W. A. Gebauer \\
	Machine Learning Group\\
	Technische Universit\"at Berlin\\
	10587 Berlin, Germany\\
	\texttt{n.wa.gebauer@gmail.com}\\
	\And
	Michael Gastegger\\
	Machine Learning Group\\
	Technische Universit\"at Berlin\\
	10587 Berlin, Germany\\
	\texttt{michael.gastegger@tu-berlin.de}\\
	\And
	Kristof T. Sch\"utt\\
	Machine Learning Group\\
	Technische Universit\"at Berlin\\
	10587 Berlin, Germany\\
	\texttt{kristof.schuett@tu-berlin.de}\\
}

\begin{document}
	
	\maketitle
	
	\begin{abstract}
		Deep learning has proven to yield fast and accurate predictions of quantum-chemical properties to accelerate the discovery of novel molecules and materials.
		As an exhaustive exploration of the vast chemical space is still infeasible, we require generative models that guide our search towards systems with desired properties.
		While graph-based models have previously been proposed, they are restricted by a lack of spatial information such that they are unable to recognize spatial isomerism and non-bonded interactions.
		Here, we introduce a generative neural network for 3d point sets that respects the rotational invariance of the targeted structures.
		We apply it to the generation of molecules and demonstrate its ability to approximate the distribution of equilibrium structures using spatial metrics as well as established measures from chemoinformatics.
		As our model is able to capture the complex relationship between 3d geometry and electronic properties, we bias the distribution of the generator towards molecules with a small HOMO-LUMO gap -- an important property for the design of organic solar cells.
	\end{abstract}
	
	\section{Introduction}
	In recent years, machine learning enabled a significant acceleration of molecular dynamics simulations~\cite{behler2007generalized,bartok2010gaussian,chmiela2017machine,gastegger2017machine,schutt2018schnet,chmiela2018towards,schutt2019unifying} as well as the prediction of chemical properties across chemical compound space~\cite{rupp2012fast,eickenberg2017solid,schutt2017quantum,gilmer2017neural,jha2018elemnet,smith2017ani}.
	Still, the discovery of novel molecules and materials with desired properties remains a major challenge in chemistry and materials science.
	Accurate geometry-based models require the atom types and positions of candidate molecules in \textit{equilibrium}, i.e. local minima of the potential energy surface.
	These have to be obtained by geometry optimization, either using computationally expensive quantum chemistry calculations or a neural network trained on both
	compositional (chemical) and configurational (structural) degrees of freedom~\cite{schutt2017schnet,smith2017ani,smith2017ani2,podryabinkin2019accelerating}.
	On the other hand, bond-based methods that predict chemical properties using only the molecular graph~\cite{duvenaud2015convolutional,ramsundar2015massively,gomez2016automatic,kearnes2016molecular} do not reach the accuracy of geometry-based methods~\cite{gilmer2017neural,wu2018moleculenet}.
	In particular, they lack crucial information to distinguish spatial isomers, i.e. molecules that exhibit the same bonding graph, but different 3d structures and chemical properties.
	As an exhaustive search of chemical space is infeasible, several methods have been proposed to generate molecular graphs~\cite{liu2018constrained,jin2018junction} and to bias the distribution of the generative model towards desired chemical properties for a guided search~\cite{you2018graph,li2018multi}.
	Naturally, these models suffer from the same drawbacks as bond-based predictions of chemical properties.
	
	In this work, we propose \textit{G-SchNet (Generative SchNet)} for the generation of rotational invariant 3d point sets.
	Points are placed sequentially according to a learned conditional probability distribution that reflects the symmetries of the previously placed points by design.
	Therefore, G-SchNet conserves rotational invariance as well as local symmetries in the predicted probabilities.
	We build upon the SchNet architecture~\cite{schutt2017schnet} using continuous-filter convolutional layers to model the interactions of already placed points.
	This allows local structures such as functional groups as well as non-bonded interactions to be captured.
	Note, that we explicitly do \emph{not} aim to generate molecular graphs, but directly the atomic types and positions, which include the full information necessary to solve the electronic problem in the Born-Oppenheimer approximation.
	This allows us to avoid abstract concepts such as bonds and rings, that are not rooted in quantum mechanics and rely heavily on heuristics.
	Here, we use these constructs only for evaluation and comparison to graph-based approaches.
	In this manner, our approach enables further analysis using atomistic simulations after generation.
	In particular, this work provides the following main contributions:
	\begin{itemize}
		\item We propose the autoregressive neural network \textit{G-SchNet} for the generation of 3d point sets incorporating the constraints of Euclidean space and rotational invariance of the atom distribution as prior knowledge\footnote{The code is publicly available at \url{www.github.com/atomistic-machine-learning/G-SchNet}}.
		\item We apply the introduced method to the generation of organic molecules with arbitrary composition based on the QM9 dataset~\cite{ramakrishnan2014quantum,reymond2015chemical,ruddigkeit2012enumeration}. 
		We show that our model yields novel and accurate equilibrium molecules while capturing spatial and structural properties of the training data distribution.
		\item We demonstrate that the generator can be biased towards complex electronic properties such as the HOMO-LUMO gap -- an important property for the design of organic solar cells.
		\item We introduce datasets containing novel, generated molecules not present in the QM9 dataset. These molecules are verified and optimized at the same level of theory used for QM9\footnote{The datasets are publicly available at \url{www.quantum-machine.org}.}.
	\end{itemize}
	
	\section{Related work}
	Existing generative models for 3d structures are usually concerned with volumetric objects represented as point cloud data~\cite{wu2016learning,achlioptas2018learning}.
	Here, the density of points characterizes shapes of the target structures.
	This involves a large amount of points where the exact placement of individual points is less important.
	Furthermore, these architectures are not designed to capture rotational invariances of the modeled objects.
	However, the accurate relative placement of points is the main requirement for the generation of molecular structures.
	
	Other generative models for molecules resort to bond-based representations instead of spatial structures. 
	They use text-processing architectures on SMILES~\cite{weininger1988smiles}~strings~(e.g.~\cite{kusner2017grammar,guimares2017objective,dai2018syntax,gomez2016automatic,janz2018learning,segler2018generating,popova2018deep,lim2018auto,blaschke2017auto,gupta2017generative,jorgensen2018machine}) or graph deep neural networks on molecular graphs~(e.g.~\cite{liu2018constrained,you2018graph,li2018multi,li2018learning,jin2018junction,jin2019learning}). 
	Although these models allow to generate novel molecular graphs, ignoring the 3d positions of atoms during generation discards valuable information connected to possible target properties.
	\citet{mansimov2019molecular} have proposed an approach to sample 3d conformers given a specific molecular graph.
	This could be combined with aforementioned graph-based generative models in order to generate 3d structures in a two-step procedure.
	For the targeted discovery of novel structures, both parts would have to be biased appropriately.
	Again, this can be problematic if the target properties crucially depend on the actual spatial configuration of molecules, as this information is lacking in the intermediate graph representation.
	
	G-SchNet, in contrast, directly generates 3d structures without relying on any graph- or bond-based information.
	It uses a spatial representation and can be trained end-to-end, which allows biasing towards complex, 3d structure-dependent electronic properties.
	Previously, we have proposed a similar factorization to generate 3d isomers given a fixed composition of atoms~\cite{gebauer2018generating}. The G-SchNet architecture improves upon this idea as it can deal with arbitrary compositions of atoms and introduces auxiliary tokens which drastically increase the scalability of the model and the robustness of the approximated distribution.
	
	\section{Symmetry-adapted factorization of point set distributions}
	We aim to factorize a distribution over point sets that enables autoregressive generation where new points are sampled step by step.
	Here, the conditional distributions for each step should be symmetry-adapted to the previously sampled points.
	This includes invariance to permutation, equivariance to rotation and translation\footnote{If we consider a point set \textit{invariant} to rotation and translation, the conditional distributions have to be \textit{equivariant} to fulfill this property.} as well as the ability to capture local symmetries, e.g. local rotations of functional groups.
	
	A 3d point set consists of a variable number $n$ of positions $\mathbf{R}_{\leq{n}}$ and types $\mathbf{Z}_{\leq n}$.
	Each position $\mathbf{r}_i \in \mathbb{R}^{3}$ is associated with a type $Z_i \in \mathbb{N}$.
	When generating molecules, these types correspond to chemical elements.
	Beyond those, we use auxiliary tokens, i.e. additional positions and types, which are not part of a generated point set but support the generation process.
	A partially generated point set including $t$ of such auxiliary tokens is denoted as $\mathbf{R}^{t}_{\leq{i}}~=~(\mathbf{r}_{1}, ...,\mathbf{r}_{t},\mathbf{r}_{t+1}, ...,\mathbf{r}_{t+i})$ and $\mathbf{Z}^{t}_{\leq i}~=~(Z_{1}, ...,Z_{t}, Z_{t+1}, ..., Z_{t+i})$, where the first $t$ indices correspond to tokens and all following positions and types indicate actual points.
	We define the distribution over a collection of point sets as the joint distribution of positions and types, $p(\mathbf{R}_{\leq{n}},\mathbf{Z}_{\leq{n}})$.
	Instead of learning the usually intractable joint distribution directly, we factorize it using conditional probabilities
	\begin{align}
	p(\mathbf{R}_{\leq{n}},\mathbf{Z}_{\leq{n}}) =
	\prod_{i=1}^{n} \Bigg[p\left(\mathbf{r}_{t+i},Z_{t+i}|\mathbf{R}^{t}_{\leq{i-1}},\mathbf{Z}^{t}_{\leq{i-1}}\right)\Bigg]
	\cdot p(\textit{stop}|\mathbf{R}^{t}_{\leq{n}},\mathbf{Z}^{t}_{\leq{n}})
	\end{align}
	that permit a sequential generation process analog to autoregressive graph generative networks~\cite{liu2018constrained,you2018graph} or PixelCNN~\cite{van2016conditional}.
	Note that the distribution over the first point in the set, $p(\mathbf{r}_{t+1}, Z_{t+1}|\mathbf{R}^{t}_{\leq{0}},\mathbf{Z}^{t}_{\leq{0}})$, is only conditioned on positions and types of the auxiliary tokens and that, depending on the tokens used, its position $r_{t+1}$ might be determined arbitrarily due to translational invariance.
	In order to allow for differently sized point sets, $p(\textit{stop}|\mathbf{R}^{t}_{\leq{n}},\mathbf{Z}^{t}_{\leq{n}})$ determines the probability of stopping the generation process.
	
	While the number and kind of tokens can be adapted to the problem at hand, there is one auxiliary token that is universally applicable and vital to the scalability of our approach: the focus point $c_i=(\mathbf{r}_{1}, Z_{1})$.
	It has its own artificial type and its position can be selected randomly at each generation step from already placed points ($\mathbf{r}_{1} \in \{\mathbf{r}_{t+1}, ..., \mathbf{r}_{t+i-1}\}$).
	We then require that the new point is in the neighborhood of $c_i$ which may be defined by a fixed radial cutoff.
	This effectively limits the potential space of new positions to a region of fixed size around $c_i$ that does not grow with the number of preceding points.
	Beyond that, we rewrite the joint distribution of type and position such that the probability of the next position $\mathbf{r}_{t+i}$ depends on its associated type $Z_{t+i}$,
	\begin{align}
	p\left(\mathbf{r}_{t+i},Z_{t+i}|\mathbf{R}^{t}_{\leq{i-1}},\mathbf{Z}^{t}_{\leq{i-1}}\right)=
	p(\mathbf{r}_{t+i}|Z_{t+i},\mathbf{R}^{t}_{\leq{i-1}},\mathbf{Z}^{t}_{\leq{i-1}}) \, p(Z_{t+i}|\mathbf{R}^{t}_{\leq{i-1}},\mathbf{Z}^{t}_{\leq{i-1}})
	\label{eq:postype}
	\end{align}
	which allows us to sample the type first and the position afterwards at each generation step.
	Note that, due to the auxiliary focus token, the sampling of types can already consider some positional information, namely that the next point will be in the neighborhood of the focus point.
	
	While the discrete distribution of types is rotationally invariant, the continuous distribution over positions transforms with already placed points.
	Instead of learning the 3d distribution of absolute positions directly, we construct it from pairwise distances to preceding points (including auxiliary tokens) which guarantees the equivariance properties:
	\begin{align}
	p(\mathbf{r}_{t+i}|\mathbf{R}^{t}_{\leq{i-1}},\mathbf{Z}^{t}_{\leq{i}}) = 
	\frac{1}{\alpha} \prod_{j=1}^{t+i-1} p(d_{(t+i)j}|\mathbf{R}^{t}_{\leq i-1},\mathbf{Z}^{t}_{\leq{i}}). \label{eq:dist_to_pos}
	\end{align}
	Here $\alpha$ is the normalization constant and $d_{(t+i)j} = ||\mathbf{r}_{t+i}-\mathbf{r}_j||_2$ is the distance between the new position and a preceding point or token.
	As the probability of $\mathbf{r}_{t+i}$ is given in these symmetry-adapted coordinates, the predicted probability is independent of translation and rotation.
	Transformed to Cartesian coordinates, this leads to rotationally equivariant probabilities of positions in 3d space according to Eq. \ref{eq:dist_to_pos}.
	Since we employ a focus point, the space to be considered at each step remains small regardless of the size of the generated point set.
	
	\section{G-SchNet}
	\begin{figure}
		\centering
		\includegraphics[width=0.8\linewidth]{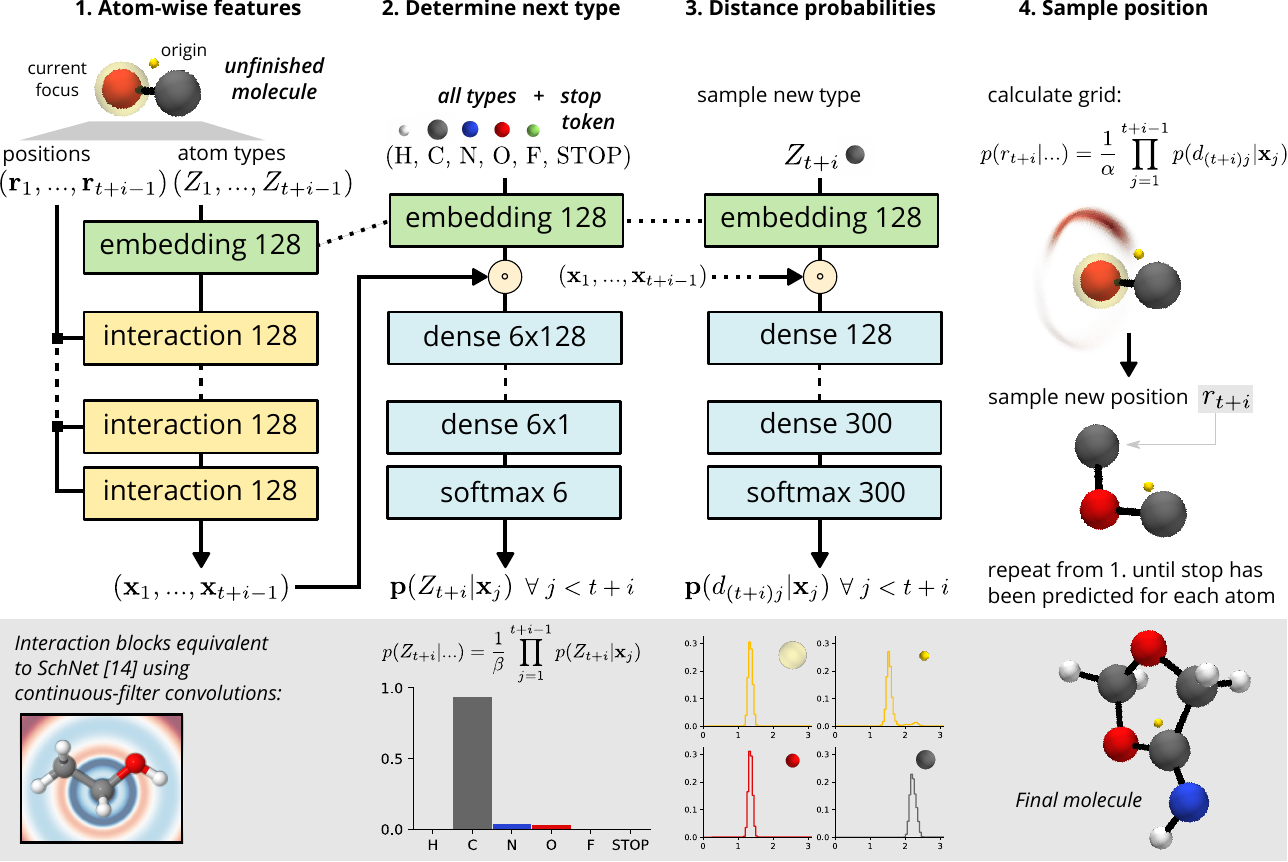}
		\caption{Scheme of G-SchNet architecture with an exemplary generation step where an atom is sampled given two already placed atoms and two auxiliary tokens: the focus and the origin. The complete molecule obtained by repeating the depicted procedure is shown in the lower right. Network layers are presented as blocks where the shape of the output is given. The embedding layers share weights and the $\odot$-operator represents element-wise multiplication of feature vectors. All molecules shown in figures have been rendered with the 3d visualization package Mayavi~\cite{ramachandran2011mayavi}.\label{fig:architecture}}
	\end{figure}
	
	\subsection{Neural network architecture}
	In order to apply the proposed generative framework to molecules, we adopt the SchNet architecture originally developed to predict quantum-chemical properties of atomistic systems~\cite{schutt2017schnet, schutt2018schnet}.
	It allows to extract atom-wise features that are invariant to rotation and translation as well as the order of atoms.
	The main building blocks are an embedding layer, mapping atom types to initial feature vectors, and interaction blocks that update the features according to the atomic environments using continuous-filter convolutions.
	We use the atom representation as a starting point to predict the necessary probability distributions outlined above.
	Fig.~\ref{fig:architecture} gives an overview of the proposed G-SchNet architecture and illustrates one step of the previously described generation process for a molecule.
	The first column shows how SchNet extracts atom-wise features from previously placed atoms and the two auxiliary tokens described below in section \ref{sec:tokens}.
	For our experiments, we use nine interaction blocks and 128 atom features.
	
	In order to predict the probability of the next type (Eq.~\ref{eq:postype}), we reuse the embedding layer of the feature extraction to embed the chemical elements as well as the stop token, which is treated as a separate atom type.
	The feature vectors are copied once for each possible next type and multiplied entry-wise with the corresponding embedding before being processed by five atom-wise dense layers with shifted softplus non-linearity\footnote{as used in SchNet~\cite{schutt2017schnet}: $\text{ssp}(x) = \ln \left(0.5 e^x + 0.5 \right)$} and a softmax.
	We obtain the final distribution as follows:
	\begin{align}
	p(Z_{t+i}|\mathbf{R}^{\textit{t}}_{\leq{i-1}},\mathbf{Z}^{t}_{\leq{i-1}}) = \frac{1}{\beta} \prod_{j=1}^{t+i-1} p\left(Z_{t+i}|\mathbf{x}_j\right).
	\end{align}
	
	The third column of Fig.~\ref{fig:architecture} shows how the distance predictions are obtained in a similar fashion where the atom features of the sampled type from column 2 are reused.
	A fully-connected network yields a distribution over distances for every input atom, which is discretized on a 1d grid.
	For our experiments, we covered distances up to $15$~{\AA} with a bin size of approximately $0.05$~{\AA}.
	The final probability distribution of positions is obtained as described in Eq.~\ref{eq:dist_to_pos}.
	We use an additional temperature parameter that allows controlling the randomness during generation.
	The exact equation and more details on all layers of the architecture are provided in the supplement.
	
	\subsection{Auxiliary tokens}\label{sec:tokens}
\begin{figure}
		\centering
		\includegraphics[width=0.7\linewidth]{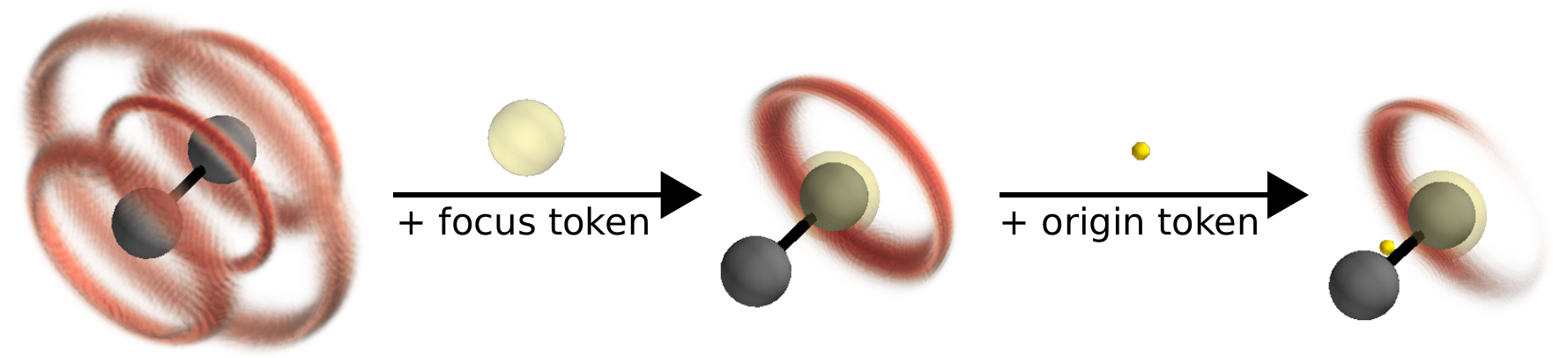}
		\caption{Effect of the introduced tokens on the distribution over positions of a new carbon atom.\label{fig:tokens}}
	\end{figure}

	We make use of two auxiliary tokens with individual, artificial types that are not part of the generated structure. 
	They are fed into the network and influence the autoregressive predictions analog to regular atoms of a molecule.
	This is illustrated in Fig.~\ref{fig:tokens}, where we show the distribution over positions of a new carbon atom given two already placed carbon atoms with and without the placement of additional tokens.
	Suppose the most probable position is next to either one of the already placed atoms, but \textit{not} between them.
	Then, without tokens, both carbon atoms predict the same distance distribution with two peaks: at a distance of one bond length and at the distance of two bond lengths to cover both likely positions of the next atom.
	This leads to high placement probabilities between the atoms as an artifact of the molecular symmetry.
	This issue can be resolved by the earlier introduced focus point $c_i$.
	Beyond localizing the prediction for scalability, the focus token breaks the symmetry as the new atom is supposed to be placed in its neighborhood.
	Thus, the generation process is guided to predict only one of the two options and the grid distribution is free of phantom positions (Fig.~\ref{fig:tokens}, middle).
	Consequently, the focus token is a vital part of G-SchNet.
	
	A second auxiliary token marks the origin around which the molecule grows.
	The origin token encodes global geometry information, as it is located at the center of mass of a molecule during training.
	In contrast to the focus token, which is placed on the currently focused atom for each placement step, the position of the origin token remains fixed during generation.
	The illustration on the right-hand side of Fig.~\ref{fig:tokens} shows how the distribution over positions is further narrowed by the origin token.
	We have carried out ablation studies where we remove the origin token from our architecture.
	These experiments are detailed in the supplement and have shown that the origin token significantly improves the approximation of the training data distribution.
	It increases the amount of validly generated structures as well as their faithfulness to typical characteristics of the training data.
	
	\subsection{Training}
	For training, we split each molecule in the training set into a trajectory of single atom placement steps.
	We sample a random atom placement trajectory in each training epoch.
	This procedure is described in detail in the supplement.
	The next type at each step $i$ can then be one-hot encoded by $\mathbf{q}_{i}^\text{type}$.
	To obtain the distance labels $\mathbf{q}_{ij}^\text{dist}$, the distances $d_{(t+i)j}$ between the preceding points (atoms and tokens) and the atom to be placed are expanded using Gaussians located at the respective bins of the 1d distance grid described above.
	The loss at step $i$ is given by the cross-entropy between labels $\mathbf{Q}_i$ and predictions $\mathbf{P}_i$
	\begin{align*}
	H\left(\mathbf{Q}_i, \mathbf{P}_i\right) = 
	\underbrace{
		\vphantom{\sum_{j=1}^i}
		-\sum_{k=1}^{n_\text{types}} \left[\mathbf{q}_{i}^\text{type}\right]_k \cdot \log\left[\mathbf{p}_{i}^\text{type}\right]_k}_
	{\textit{cross-entropy of types}}
	\underbrace{
		-\frac{1}{t+i-1}\sum_{j=1}^{t+i-1}\sum_{l=1}^{n_\text{bins}} \left[\mathbf{q}_{ij}^\text{dist}\right]_l \cdot \log\left[\mathbf{p}_{ij}^\text{dist}\right]_l}_
	{\textit{average cross-entropy of distances}},
	\end{align*}
	where $\mathbf{p}_{i}^\text{type} = \frac{1}{\beta} \prod_{j=1}^{t+i-1} \mathbf{p}(Z_{t+i}|\mathbf{x}_j)$ with normalizing constant $\beta$ is the predicted type distribution and $\mathbf{p}_{ij}^\text{dist} = \mathbf{p}(d_{(t+i)j}|\mathbf{x}_j)$ is the predicted distance distribution.
	For further details on the training procedure, please refer to the supplement.
	
	\section{Experiments and results}
	We train G-SchNet on a randomly selected subset of 50k molecules from the QM9 benchmark dataset~\cite{ramakrishnan2014quantum,reymond2015chemical,ruddigkeit2012enumeration} consisting of \textasciitilde134k organic molecules with up to nine heavy atoms from carbon, nitrogen, oxygen, and fluorine.
	We use 5k molecules for validation and early stopping while the remaining data is used as a test set.
	All models are trained with the ADAM optimizer~\cite{kingma2014adam}.
	To implement G-SchNet, we build upon the SchNetPack framework~\cite{schutt2019schnetpack} which uses PyTorch~\cite{pytorch2019neurips} and ASE~\cite{larsen2017ase}.
	
	After training, we generate 20k molecules.
	In our analysis, we filter generated molecules for \textit{validity}, i.e. we only consider structures that have no disconnected parts and where the number of bonds is equal to the valence of the respective atom type for all atoms.
	We use Open Babel~\cite{openbabel} to obtain (kekulized) bond orders of generated structures for the valency check.
	In our experiments, about 77\% of generated molecules are valid.
	For the statistics presented in the following, we also filter non-unique structures removing approximately 10\%.
	A step-by-step description of the generation process, details on the subsequent matching of molecules, and tables with extended statistics of G-SchNet and related generative models are given in the supplement.
	
	\subsection{Accuracy of generated molecules}
	\label{sec:accuracy}
\begin{figure}
		\centering
		\includegraphics[width=0.95\linewidth]{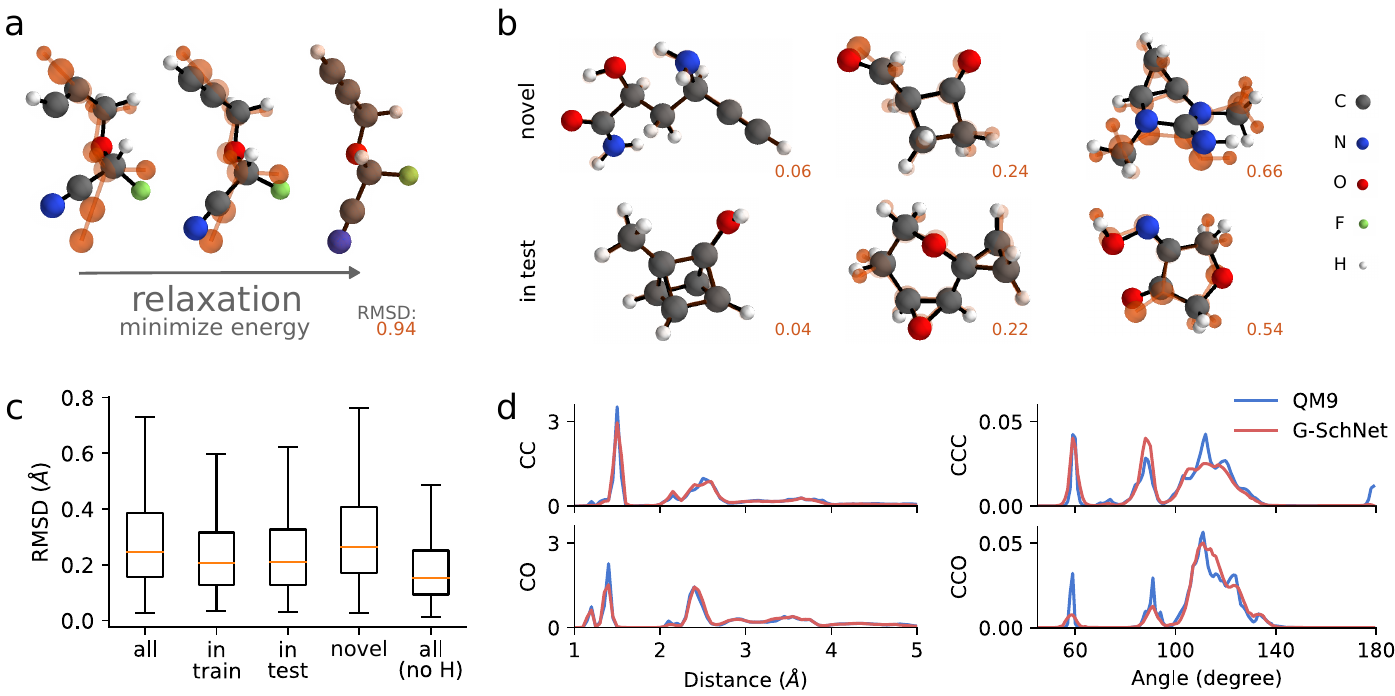}
		\caption{(a)~Scheme showing the relaxation of a molecule along with the resulting root-mean-square deviation (RMSD) between its atomic positions before and after relaxation. The equilibrium structure is indicated by orange shadows. (b)~Examples of generated molecules which are novel (top) or resemble test data (bottom). The RMSD increases from left to right. As before, the orange shadows indicate the corresponding equilibrium configurations. (c)~Boxplots of the RMSD for generated molecules divided into different sets. The boxes extend from the lower to the upper quartile and the whiskers reach up to 1.5 times the interquartile range. (d)~Radial distribution functions for carbon-carbon and carbon-oxygen atom pairs (left) and angular distribution functions for bonded carbon-carbon-carbon and carbon-carbon-oxygen chains (right) in the training data and in generated molecules.\label{fig:rmsd}}
	\end{figure}

	Unlike molecular graphs which implicitly assume a structural equilibrium, G-SchNet is not restricted to a particular configuration and instead learns to generate equilibrium molecules purely by training on the QM9 data set.
	In  order to assess the quality of our model, we need to compare the generated 3d structures to their relaxed counterparts, i.e. the closest local minimum on the potential energy surface.
	The latter are found by minimizing the energy with respect to the atom positions at the same level of theory that was used for the QM9 dataset (B3LYP/6-31G(2df,p)~\cite{Becke1993JCP,Lee1988PRBb,Vosko1980CJP,Stephens1994JPC} with the Gaussian version of B3LYP) using the Orca quantum chemistry package~\cite{Neese2012WCMS}.
	Afterwards, we calculate the root-mean-square deviation (RMSD) of atomic positions between generated and relaxed, ground-truth geometries.
	The relaxation process is shown in Fig.~\ref{fig:rmsd}a for an example molecule.
	
	Fig.~\ref{fig:rmsd}b shows examples of \textit{unseen} generated molecules that are not in the training set with low, average, and high RMSD (left to right).
	The molecules in the middle column with an RMSD around 0.2~{\AA} closely match the equilibrium configuration.
	Even the examples with high RMSD do not change excessively.
	In fact, most of the generated molecules are close to equilibrium as the error distribution shown in Fig.~\ref{fig:rmsd}c reveals.
	It shows boxplots for molecules which resemble training data (\textit{in train}), unseen test data (\textit{in test}), and \textit{novel} molecules that cannot be found in QM9.
	We observe that unseen test structures are generated as accurately as training examples with both exhibiting a median RMSD around 0.21~{\AA}.
	The test error reported by \citet{mansimov2019molecular} for their molecular graph to 3d structure translation model is almost twice as high (0.39~{\AA} without and 0.37~{\AA} with force field post-processing). 
	However, these results are not directly comparable as they do not relax generated structures but measure the median of the mean RMSD between 100 predicted conformers per graph and the corresponding ground-truth configuration from QM9.
	For the novel geometries, the RMSDs depicted in the boxplot are slightly higher than in the training or test data cases.
	This is most likely caused by large molecules with more than nine heavy atoms, which make up almost one-third of the novel structures.
	Overall the RMSDs are still sufficiently small to conclude that novel molecules are generated close to equilibrium.
	The last boxplot shows RMSDs considering only heavy atoms of all generated molecules.
	It is significantly lower than the RMSDs of all atoms, which suggests that the necessary relaxation is to a large part concerned with the rearrangement of hydrogens.
	
	To judge how well our model fits the distribution of QM9, we compare the radial distribution functions of the two most common heavy atom pairs (carbon-carbon and carbon-oxygen) in Fig.~\ref{fig:rmsd}d.
	They align well as our training objective directly penalizes deviations in the prediction of distances.
	Beyond that we plotted angular distribution functions which are not directly related to the optimization objective. 
	They show the distribution over angles of bonded carbon-carbon-carbon and carbon-carbon-oxygen chains.
	While our model produces smoother peaks here, they also align well indicating that the angles between atoms are accurately reproduced. 
	This is remarkable insofar that only distances have been used to calculate the probabilities of positions.
	
	\subsection{Structural properties of generated molecules}
	As proposed by \citet{liu2018constrained}, we calculate the atom, bond, and ring counts of the generated molecules to see if G-SchNet captures these structural statistics of the training data.
	We use the same sets of valid and unique molecules as for the spatial analysis above.
	The atom and bond counts are determined with Open Babel.
	For ring counts, the canonical SMILES representation of generated molecules is read with RDKit~\cite{rdkit} and the symmetrized smallest set of smallest rings is computed.
	
	We compute the statistics for QM9, molecules generated with G-SchNet as well as molecules obtained by the CGVAE model of \citet{liu2018constrained}, a graph-based generative model for molecules (structures provided by the authors).
	The CGVAE model incorporates valency constraints in its generation process, allowing it to sample only valid molecules.
	This cannot easily be transferred to 3d models since there is no direct correspondence to bond order for some spatial arrangements.
	This is because the associated bond order may depend on the not yet sampled atomic environment.
	However, the valency check is a comparably fast post-processing step that allows us to remove invalid molecules at low cost.
	Furthermore, in contrast to our approach, CGVAE does not actively place hydrogen atoms and uses all available structures from QM9 for training.
	\begin{figure}
		\centering
		\includegraphics[width=0.9\linewidth]{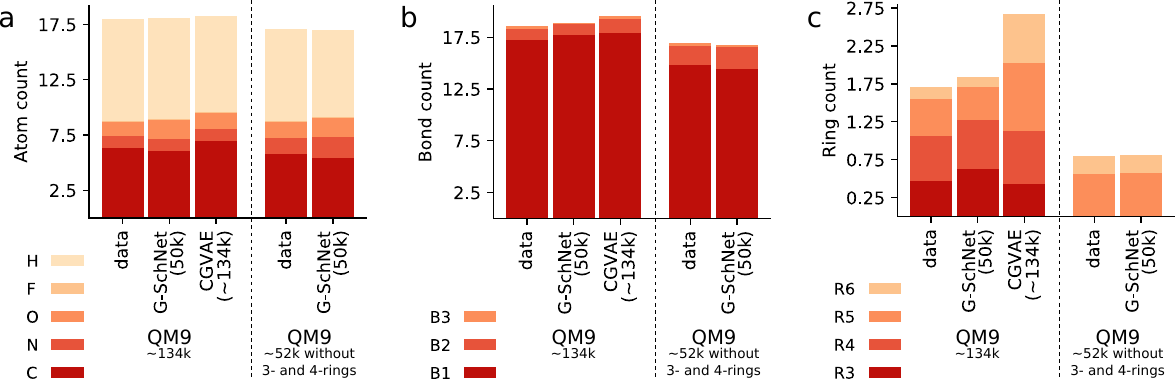}
		\caption{Bar plots showing the average numbers of atoms, bonds, and rings per molecule. Left-hand side of plots: QM9 dataset, generated molecules of G-SchNet, and generated molecules of CGVAE~\cite{liu2018constrained}. Right-hand side of plots: QM9 subset without 3- and 4-rings and generated molecules of G-SchNet trained on that subset. Numbers below model names represent the amount of training data used. B1, B2, and B3 correspond to single, double, and triple bonds. R3, R4, R5, R6 are rings of size 3 to 6.\label{fig:stats}}
	\end{figure}
	
	Fig.~\ref{fig:stats} shows the statistics for the QM9 dataset (first bars of each plot), G-SchNet (second bars), as well as CGVAE (third bars). 
	We observe that G-SchNet captures the atom and bond count of the dataset accurately.
	CGVAE shows comparable results with a slightly increased heavy atom and bond count.
	The ring count of CGVAE is significantly higher than in the training data distribution, especially concerning five- and six-membered rings.
	The average number of rings per molecule is better reproduced by G-SchNet, but here we observe an increase in three- and four-membered rings.
	
	As illustrated in Fig.~\ref{fig:tokens}, approximating positions in terms of distances can lead to symmetry artifacts at wrong bond angles if no auxiliary tokens are used.
	In order to rule out that the increase of three- and four-membered rings is an artifact of our model, we filter all molecules containing such rings from QM9.
	We train G-SchNet on the resulting subset of approximately 52k structures, where we use 50k molecules for training and the remaining 2k for validation.
	If our model was biased towards small rings, we expect to find them in the generated molecules regardless.
	We generate 20k molecules with this model and keep structures that are both valid (77\% of all generated) and unique (89\% of all valid).
	The right-hand side of each bar plot in Fig.~\ref{fig:stats} shows the corresponding results, comparing the statistics of the QM9 subset (fourth bars) with those of the generated molecules (fifth bars).
	Again, the atom and bond count statistics are accurately matched.
	More importantly, the generated molecules almost perfectly resemble the ring count of the training data, exhibiting no three- or four-membered rings.
	This hints at the possibility to bias the predicted distribution by constraining the training data.
	
	\subsection{Targeted discovery of molecules with small HOMO-LUMO gaps}
	Finally, we aim to guide the generator towards a desired value range of a complex electronic property, namely a small HOMO-LUMO gap.
	This property describes the energy difference between the highest occupied and lowest unoccupied molecular orbital and is an important measure for designing organic semiconductors, an essential component of e.g. organic solar cells and OLED displays.
	To this end, we filter all \textasciitilde3.8k equilibrium structures from QM9 with a HOMO-LUMO gap smaller than 4.5~eV.
	Then we fine-tune the G-SchNet model previously trained on all 50k molecules of the training set on this small subset.
	We use 3.3k structures for training and the remaining 0.5k for validation and early stopping.
	The optimizer is reset to its initial learning rate and the optimization protocol for biasing is the same as for regular training.
	
	We generate 20k molecules with this fine-tuned model.
	The validity decreases to 69\% of which 74\% are unique.
	This drop in validity and uniqueness is expected as the amount of generated molecules is significantly higher than the amount of available training structures with the target property.
	We relax the valid and unique molecules using the same level of theory as described in section~\ref{sec:accuracy}.
	\begin{figure}
		\centering
		\includegraphics[width=0.9\linewidth]{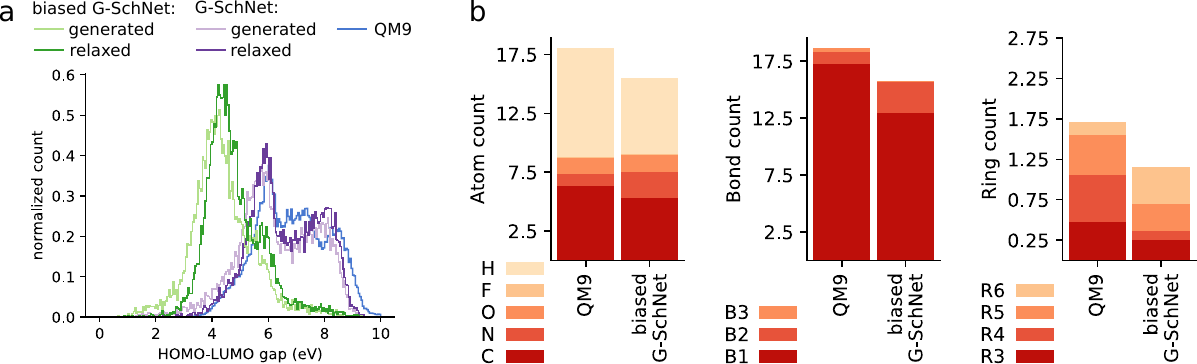}
		\caption{(a)~Histograms of calculated HOMO-LUMO gaps for molecules generated with the biased G-SchNet (green curves), G-SchNet before biasing (purple curves), and for the QM9 dataset. (b)~Bar plots showing the average numbers of atoms, bonds, and rings per molecule for QM9 and for molecules generated with the biased G-SchNet. B1, B2, and B3 correspond to single, double, and triple bonds. R3, R4, R5, R6 are rings of size 3 to 6.\label{fig:gap_results}}
	\end{figure}
	
	Fig.~\ref{fig:gap_results}a shows histograms of calculated HOMO-LUMO gaps for generated molecules before and after relaxation.
	There is a significant shift to smaller values compared to the histograms of HOMO-LUMO gaps of the full QM9 dataset.
	Before biasing, only 7\% of all validly generated molecules had a HOMO-LUMO gap smaller than 5.0~eV.
	The fine-tuning raises this number to 43\%.
	We observe in Fig.~\ref{fig:gap_results}b how the structural statistics of molecules generated with the biased G-SchNet deviate from the statistics of all molecules in QM9.
	Compared to the original dataset, the new molecules exhibit an increased number of double bonds in addition to a tendency towards forming six-membered cycles.
	These features indicate the presence of conjugated systems of alternating single and double bonds, as well as aromatic rings, both of which are important motifs in organic semiconductors.
	
	\section{Conclusions}
	We have developed the autoregressive G-SchNet for the symmetry-adapted generation of rotationally invariant 3d point sets.
	In contrast to graph generative networks and previous work on point cloud generation, our model both incorporates the constraints of euclidean space and the spatial invariances of the targeted geometries.
	This is achieved by determining the next position using distances to previously placed points, resulting in an equivariant conditional probability distribution.
	Beyond these invariances, local symmetries are captured by means of point-wise features from the SchNet architecture for atomistic predictions.
	
	We have shown that G-SchNet is able to generate accurate 3d equilibrium structures of organic molecules.
	The generated molecules closely resemble both spatial and structural distributions of the training data while our model yields 79\% unseen molecules.
	On this basis, we have introduced a dataset of >9k novel organic molecules that are not contained in QM9.
	Finally, we fine-tuned G-SchNet on a small subset of molecules to bias the distribution of the generator towards molecules with small HOMO-LUMO gap -- an important electronic property for the design of organic solar cells.
	We have generated a dataset of >3.6k novel structures with the desired property, demonstrating a promising strategy for targeted discovery of molecules.
	Further experiments documented in the supplementary material demonstrate that G-SchNet can be biased towards other electronic properties in the same manner.
	
	Directions for future work include scaling to larger systems, a direct conditioning on chemical properties as well as including periodic boundary conditions.
	We expect G-SchNet to be widely applicable, for instance, to the guided exploration of chemical space or crystal structure prediction.
	
	\subsubsection*{Acknowledgments}
	We thank \citet{liu2018constrained} for providing us with generated molecules of their CGVAE model.
	This work was supported by the Federal Ministry of Education and Research (BMBF) for the Berlin Center for Machine Learning (01IS18037A).
	MG was provided financial support by the European Unions Horizon 2020 research and innovation program under the Marie Sk\l{}odowska-Curie grant agreement NO 792572. Correspondence to NWAG and KTS.
	
	\clearpage
	
	\bibliographystyle{abbr_unsrtnat}
	\bibliography{references.bib}

\begin{thebibliography}{65}
\providecommand{\natexlab}[1]{#1}
\providecommand{\url}[1]{\texttt{#1}}
\expandafter\ifx\csname urlstyle\endcsname\relax
  \providecommand{\doi}[1]{doi: #1}\else
  \providecommand{\doi}{doi: \begingroup \urlstyle{rm}\Url}\fi

\bibitem[Behler and Parrinello(2007)]{behler2007generalized}
J.~Behler and M.~Parrinello.
\newblock Generalized neural-network representation of high-dimensional
  potential-energy surfaces.
\newblock \emph{Phys. Rev. Lett.}, 98\penalty0 (14):\penalty0 146401, 2007.

\bibitem[Bart{\'o}k et~al.(2010)Bart{\'o}k, Payne, Kondor, and
  Cs{\'a}nyi]{bartok2010gaussian}
A.~P. Bart{\'o}k, M.~C. Payne, R.~Kondor, and G.~Cs{\'a}nyi.
\newblock Gaussian approximation potentials: The accuracy of quantum mechanics,
  without the electrons.
\newblock \emph{Phys. Rev. Lett.}, 104\penalty0 (13):\penalty0 136403, 2010.

\bibitem[Chmiela et~al.(2017)Chmiela, Tkatchenko, Sauceda, Poltavsky,
  Sch{\"u}tt, and M{\"u}ller]{chmiela2017machine}
S.~Chmiela, A.~Tkatchenko, H.~E. Sauceda, I.~Poltavsky, K.~T. Sch{\"u}tt, and
  K.-R. M{\"u}ller.
\newblock Machine learning of accurate energy-conserving molecular force
  fields.
\newblock \emph{Sci. Adv.}, 3\penalty0 (5):\penalty0 e1603015, 2017.

\bibitem[Gastegger et~al.(2017)Gastegger, Behler, and
  Marquetand]{gastegger2017machine}
M.~Gastegger, J.~Behler, and P.~Marquetand.
\newblock Machine learning molecular dynamics for the simulation of infrared
  spectra.
\newblock \emph{Chem. Sci.}, 8\penalty0 (10):\penalty0 6924--6935, 2017.

\bibitem[Sch{\"u}tt et~al.(2018)Sch{\"u}tt, Sauceda, Kindermans, Tkatchenko,
  and M{\"u}ller]{schutt2018schnet}
K.~T. Sch{\"u}tt, H.~E. Sauceda, P.-J. Kindermans, A.~Tkatchenko, and K.-R.
  M{\"u}ller.
\newblock {SchNet -- A} deep learning architecture for molecules and materials.
\newblock \emph{The Journal of Chemical Physics}, 148\penalty0 (24):\penalty0
  241722, 2018.

\bibitem[Chmiela et~al.(2018)Chmiela, Sauceda, M{\"u}ller, and
  Tkatchenko]{chmiela2018towards}
S.~Chmiela, H.~E. Sauceda, K.-R. M{\"u}ller, and A.~Tkatchenko.
\newblock Towards exact molecular dynamics simulations with machine-learned
  force fields.
\newblock \emph{Nature communications}, 9\penalty0 (1):\penalty0 3887, 2018.

\bibitem[Sch{\"u}tt et~al.(2019)Sch{\"u}tt, Gastegger, Tkatchenko, M{\"u}ller,
  and Maurer]{schutt2019unifying}
K.~Sch{\"u}tt, M.~Gastegger, A.~Tkatchenko, K.-R. M{\"u}ller, and R.~J. Maurer.
\newblock Unifying machine learning and quantum chemistry with a deep neural
  network for molecular wavefunctions.
\newblock \emph{Nature communications}, 10\penalty0 (1):\penalty0 5024, 2019.

\bibitem[Rupp et~al.(2012)Rupp, Tkatchenko, M{\"u}ller, and
  Von~Lilienfeld]{rupp2012fast}
M.~Rupp, A.~Tkatchenko, K.-R. M{\"u}ller, and O.~A. Von~Lilienfeld.
\newblock Fast and accurate modeling of molecular atomization energies with
  machine learning.
\newblock \emph{Phys. Rev. Lett.}, 108\penalty0 (5):\penalty0 058301, 2012.

\bibitem[Eickenberg et~al.(2017)Eickenberg, Exarchakis, Hirn, and
  Mallat]{eickenberg2017solid}
M.~Eickenberg, G.~Exarchakis, M.~Hirn, and S.~Mallat.
\newblock Solid harmonic wavelet scattering: Predicting quantum molecular
  energy from invariant descriptors of 3d electronic densities.
\newblock In \emph{Advances in Neural Information Processing Systems 30}, pages
  6543--6552. Curran Associates, Inc., 2017.

\bibitem[Sch{\"u}tt et~al.(2017{\natexlab{a}})Sch{\"u}tt, Arbabzadah, Chmiela,
  M{\"u}ller, and Tkatchenko]{schutt2017quantum}
K.~T. Sch{\"u}tt, F.~Arbabzadah, S.~Chmiela, K.~R. M{\"u}ller, and
  A.~Tkatchenko.
\newblock Quantum-chemical insights from deep tensor neural networks.
\newblock \emph{Nature Communications}, 8:\penalty0 13890, 2017{\natexlab{a}}.

\bibitem[Gilmer et~al.(2017)Gilmer, Schoenholz, Riley, Vinyals, and
  Dahl]{gilmer2017neural}
J.~Gilmer, S.~S. Schoenholz, P.~F. Riley, O.~Vinyals, and G.~E. Dahl.
\newblock Neural message passing for quantum chemistry.
\newblock In \emph{Proceedings of the 34th International Conference on Machine
  Learning}, pages 1263--1272, 2017.

\bibitem[Jha et~al.(2018)Jha, Ward, Paul, Liao, Choudhary, Wolverton, and
  Agrawal]{jha2018elemnet}
D.~Jha, L.~Ward, A.~Paul, W.-k. Liao, A.~Choudhary, C.~Wolverton, and
  A.~Agrawal.
\newblock Elemnet: Deep learning the chemistry of materials from only elemental
  composition.
\newblock \emph{Scientific reports}, 8\penalty0 (1):\penalty0 17593, 2018.

\bibitem[Smith et~al.(2017{\natexlab{a}})Smith, Isayev, and
  Roitberg]{smith2017ani}
J.~S. Smith, O.~Isayev, and A.~E. Roitberg.
\newblock Ani-1: an extensible neural network potential with dft accuracy at
  force field computational cost.
\newblock \emph{Chem. Sci.}, 8\penalty0 (4):\penalty0 3192--3203,
  2017{\natexlab{a}}.

\bibitem[Sch{\"u}tt et~al.(2017{\natexlab{b}})Sch{\"u}tt, Kindermans, Felix,
  Chmiela, Tkatchenko, and M{\"u}ller]{schutt2017schnet}
K.~Sch{\"u}tt, P.-J. Kindermans, H.~E.~S. Felix, S.~Chmiela, A.~Tkatchenko, and
  K.-R. M{\"u}ller.
\newblock {SchNet}: A continuous-filter convolutional neural network for
  modeling quantum interactions.
\newblock In \emph{Advances in Neural Information Processing Systems}, pages
  992--1002, 2017{\natexlab{b}}.

\bibitem[Smith et~al.(2017{\natexlab{b}})Smith, Isayev, and
  Roitberg]{smith2017ani2}
J.~S. Smith, O.~Isayev, and A.~E. Roitberg.
\newblock Ani-1, a data set of 20 million calculated off-equilibrium
  conformations for organic molecules.
\newblock \emph{Scientific data}, 4:\penalty0 170193, 2017{\natexlab{b}}.

\bibitem[Podryabinkin et~al.(2019)Podryabinkin, Tikhonov, Shapeev, and
  Oganov]{podryabinkin2019accelerating}
E.~V. Podryabinkin, E.~V. Tikhonov, A.~V. Shapeev, and A.~R. Oganov.
\newblock Accelerating crystal structure prediction by machine-learning
  interatomic potentials with active learning.
\newblock \emph{Physical Review B}, 99\penalty0 (6):\penalty0 064114, 2019.

\bibitem[Duvenaud et~al.(2015)Duvenaud, Maclaurin, Iparraguirre, Bombarell,
  Hirzel, Aspuru-Guzik, and Adams]{duvenaud2015convolutional}
D.~K. Duvenaud, D.~Maclaurin, J.~Iparraguirre, R.~Bombarell, T.~Hirzel,
  A.~Aspuru-Guzik, and R.~P. Adams.
\newblock Convolutional networks on graphs for learning molecular fingerprints.
\newblock In C.~Cortes, N.~D. Lawrence, D.~D. Lee, M.~Sugiyama, and R.~Garnett,
  editors, \emph{NIPS}, pages 2224--2232, 2015.

\bibitem[Ramsundar et~al.(2015)Ramsundar, Kearnes, Riley, Webster, Konerding,
  and Pande]{ramsundar2015massively}
B.~Ramsundar, S.~Kearnes, P.~Riley, D.~Webster, D.~Konerding, and V.~Pande.
\newblock Massively multitask networks for drug discovery.
\newblock \emph{arXiv preprint arXiv:1502.02072}, 2015.

\bibitem[G{\'o}mez-Bombarelli et~al.(2016)G{\'o}mez-Bombarelli, Wei, Duvenaud,
  Hern{\'a}ndez-Lobato, S{\'a}nchez-Lengeling, Sheberla, Aguilera-Iparraguirre,
  Hirzel, Adams, and Aspuru-Guzik]{gomez2016automatic}
R.~G{\'o}mez-Bombarelli, J.~N. Wei, D.~Duvenaud, J.~M. Hern{\'a}ndez-Lobato,
  B.~S{\'a}nchez-Lengeling, D.~Sheberla, J.~Aguilera-Iparraguirre, T.~D.
  Hirzel, R.~P. Adams, and A.~Aspuru-Guzik.
\newblock Automatic chemical design using a data-driven continuous
  representation of molecules.
\newblock \emph{ACS Cent. Sci.}, 2016.

\bibitem[Kearnes et~al.(2016)Kearnes, McCloskey, Berndl, Pande, and
  Riley]{kearnes2016molecular}
S.~Kearnes, K.~McCloskey, M.~Berndl, V.~Pande, and P.~Riley.
\newblock Molecular graph convolutions: moving beyond fingerprints.
\newblock \emph{J. Comput. Aided Mol. Des.}, 30\penalty0 (8):\penalty0
  595--608, 2016.

\bibitem[Wu et~al.(2018)Wu, Ramsundar, Feinberg, Gomes, Geniesse, Pappu,
  Leswing, and Pande]{wu2018moleculenet}
Z.~Wu, B.~Ramsundar, E.~N. Feinberg, J.~Gomes, C.~Geniesse, A.~S. Pappu,
  K.~Leswing, and V.~Pande.
\newblock Moleculenet: a benchmark for molecular machine learning.
\newblock \emph{Chemical science}, 9\penalty0 (2):\penalty0 513--530, 2018.

\bibitem[Liu et~al.(2018)Liu, Allamanis, Brockschmidt, and
  Gaunt]{liu2018constrained}
Q.~Liu, M.~Allamanis, M.~Brockschmidt, and A.~Gaunt.
\newblock Constrained graph variational autoencoders for molecule design.
\newblock In S.~Bengio, H.~Wallach, H.~Larochelle, K.~Grauman, N.~Cesa-Bianchi,
  and R.~Garnett, editors, \emph{Advances in Neural Information Processing
  Systems 31}, pages 7795--7804. Curran Associates, Inc., 2018.

\bibitem[Jin et~al.(2018)Jin, Barzilay, and Jaakkola]{jin2018junction}
W.~Jin, R.~Barzilay, and T.~Jaakkola.
\newblock Junction tree variational autoencoder for molecular graph generation.
\newblock \emph{arXiv preprint arXiv:1802.04364}, 2018.

\bibitem[You et~al.(2018)You, Liu, Ying, Pande, and Leskovec]{you2018graph}
J.~You, B.~Liu, Z.~Ying, V.~Pande, and J.~Leskovec.
\newblock Graph convolutional policy network for goal-directed molecular graph
  generation.
\newblock In S.~Bengio, H.~Wallach, H.~Larochelle, K.~Grauman, N.~Cesa-Bianchi,
  and R.~Garnett, editors, \emph{Advances in Neural Information Processing
  Systems 31}, pages 6410--6421. Curran Associates, Inc., 2018.

\bibitem[Li et~al.(2018{\natexlab{a}})Li, Zhang, and Liu]{li2018multi}
Y.~Li, L.~Zhang, and Z.~Liu.
\newblock Multi-objective de novo drug design with conditional graph generative
  model.
\newblock \emph{Journal of cheminformatics}, 10\penalty0 (1):\penalty0 33,
  2018{\natexlab{a}}.

\bibitem[Ramakrishnan et~al.(2014)Ramakrishnan, Dral, Rupp, and von
  Lilienfeld]{ramakrishnan2014quantum}
R.~Ramakrishnan, P.~O. Dral, M.~Rupp, and O.~A. von Lilienfeld.
\newblock Quantum chemistry structures and properties of 134 kilo molecules.
\newblock \emph{Sci. Data}, 1, 2014.

\bibitem[Reymond(2015)]{reymond2015chemical}
J.-L. Reymond.
\newblock The chemical space project.
\newblock \emph{Acc. Chem. Res.}, 48\penalty0 (3):\penalty0 722--730, 2015.

\bibitem[Ruddigkeit et~al.(2012)Ruddigkeit, Van~Deursen, Blum, and
  Reymond]{ruddigkeit2012enumeration}
L.~Ruddigkeit, R.~Van~Deursen, L.~C. Blum, and J.-L. Reymond.
\newblock Enumeration of 166 billion organic small molecules in the chemical
  universe database {GDB-17}.
\newblock \emph{Journal of chemical information and modeling}, 52\penalty0
  (11):\penalty0 2864--2875, 2012.

\bibitem[Wu et~al.(2016)Wu, Zhang, Xue, Freeman, and Tenenbaum]{wu2016learning}
J.~Wu, C.~Zhang, T.~Xue, B.~Freeman, and J.~Tenenbaum.
\newblock Learning a probabilistic latent space of object shapes via 3d
  generative-adversarial modeling.
\newblock In D.~D. Lee, M.~Sugiyama, U.~V. Luxburg, I.~Guyon, and R.~Garnett,
  editors, \emph{Advances in Neural Information Processing Systems 29}, pages
  82--90. Curran Associates, Inc., 2016.

\bibitem[Achlioptas et~al.(2018)Achlioptas, Diamanti, Mitliagkas, and
  Guibas]{achlioptas2018learning}
P.~Achlioptas, O.~Diamanti, I.~Mitliagkas, and L.~Guibas.
\newblock Learning representations and generative models for 3d point clouds.
\newblock In \emph{International Conference on Machine Learning}, pages 40--49,
  2018.

\bibitem[Weininger(1988)]{weininger1988smiles}
D.~Weininger.
\newblock {SMILES}, a chemical language and information system. 1.
  {I}ntroduction to methodology and encoding rules.
\newblock \emph{Journal of Chemical Information and Computer Sciences},
  28\penalty0 (1):\penalty0 31--36, 1988.

\bibitem[Kusner et~al.(2017)Kusner, Paige, and
  Hern{\'a}ndez-Lobato]{kusner2017grammar}
M.~J. Kusner, B.~Paige, and J.~M. Hern{\'a}ndez-Lobato.
\newblock Grammar variational autoencoder.
\newblock In \emph{Proceedings of the 34th International Conference on Machine
  Learning-Volume 70}, pages 1945--1954. JMLR. org, 2017.

\bibitem[Guimaraes et~al.(2017)Guimaraes, Sanchez-Lengeling, Farias, and
  Aspuru-Guzik]{guimares2017objective}
G.~L. Guimaraes, B.~Sanchez-Lengeling, P.~L.~C. Farias, and A.~Aspuru-Guzik.
\newblock Objective-reinforced generative adversarial networks ({ORGAN}) for
  sequence generation models.
\newblock \emph{arXiv preprint arXiv:1705.10843}, 2017.

\bibitem[Dai et~al.(2018)Dai, Tian, Dai, Skiena, and Song]{dai2018syntax}
H.~Dai, Y.~Tian, B.~Dai, S.~Skiena, and L.~Song.
\newblock Syntax-directed variational autoencoder for structured data.
\newblock In \emph{International Conference on Learning Representations}, 2018.

\bibitem[Janz et~al.(2018)Janz, van~der Westhuizen, Paige, Kusner, and
  Lobato]{janz2018learning}
D.~Janz, J.~van~der Westhuizen, B.~Paige, M.~Kusner, and J.~M.~H. Lobato.
\newblock Learning a generative model for validity in complex discrete
  structures.
\newblock In \emph{International Conference on Learning Representations}, 2018.

\bibitem[Segler et~al.(2018)Segler, Kogej, Tyrchan, and
  Waller]{segler2018generating}
M.~H.~S. Segler, T.~Kogej, C.~Tyrchan, and M.~P. Waller.
\newblock Generating focused molecule libraries for drug discovery with
  recurrent neural networks.
\newblock \emph{ACS Central Science}, 4\penalty0 (1):\penalty0 120--131, 2018.

\bibitem[Popova et~al.(2018)Popova, Isayev, and Tropsha]{popova2018deep}
M.~Popova, O.~Isayev, and A.~Tropsha.
\newblock Deep reinforcement learning for de novo drug design.
\newblock \emph{Science Advances}, 4\penalty0 (7), 2018.

\bibitem[Lim et~al.(2018)Lim, Ryu, Kim, and Kim]{lim2018auto}
J.~Lim, S.~Ryu, J.~W. Kim, and W.~Y. Kim.
\newblock Molecular generative model based on conditional variational
  autoencoder for de novo molecular design.
\newblock In \emph{J. Cheminformatics}, 2018.

\bibitem[Blaschke et~al.(2017)Blaschke, Olivecrona, Engkvist, Bajorath, and
  Chen]{blaschke2017auto}
T.~Blaschke, M.~Olivecrona, O.~Engkvist, J.~Bajorath, and H.~Chen.
\newblock Application of generative autoencoder in de novo molecular design.
\newblock \emph{Molecular Informatics}, 37\penalty0 (1-2):\penalty0 1700123,
  2017.

\bibitem[Gupta et~al.(2017)Gupta, Müller, Huisman, Fuchs, Schneider, and
  Schneider]{gupta2017generative}
A.~Gupta, A.~T. Müller, B.~J.~H. Huisman, J.~A. Fuchs, P.~Schneider, and
  G.~Schneider.
\newblock Generative recurrent networks for de novo drug design.
\newblock \emph{Molecular Informatics}, 37\penalty0 (1-2):\penalty0 1700111,
  2017.

\bibitem[J{\o}rgensen et~al.(2018)J{\o}rgensen, Mesta, Shil,
  Garc{\'\i}a~Lastra, Jacobsen, Thygesen, and Schmidt]{jorgensen2018machine}
P.~B. J{\o}rgensen, M.~Mesta, S.~Shil, J.~M. Garc{\'\i}a~Lastra, K.~W.
  Jacobsen, K.~S. Thygesen, and M.~N. Schmidt.
\newblock Machine learning-based screening of complex molecules for polymer
  solar cells.
\newblock \emph{The Journal of Chemical Physics}, 148\penalty0 (24):\penalty0
  241735, 2018.

\bibitem[Li et~al.(2018{\natexlab{b}})Li, Vinyals, Dyer, Pascanu, and
  Battaglia]{li2018learning}
Y.~Li, O.~Vinyals, C.~Dyer, R.~Pascanu, and P.~Battaglia.
\newblock Learning deep generative models of graphs.
\newblock \emph{arXiv preprint arXiv:1803.03324}, 2018{\natexlab{b}}.

\bibitem[Jin et~al.(2019)Jin, Yang, Barzilay, and Jaakkola]{jin2019learning}
W.~Jin, K.~Yang, R.~Barzilay, and T.~Jaakkola.
\newblock Learning multimodal graph-to-graph translation for molecular
  optimization.
\newblock In \emph{International Conference on Learning Representations}, 2019.

\bibitem[Mansimov et~al.(2019)Mansimov, Mahmood, Kang, and
  Cho]{mansimov2019molecular}
E.~Mansimov, O.~Mahmood, S.~Kang, and K.~Cho.
\newblock Molecular geometry prediction using a deep generative graph neural
  network.
\newblock \emph{arXiv preprint arXiv:1904.00314}, 2019.

\bibitem[Gebauer et~al.(2018)Gebauer, Gastegger, and
  Sch{\"u}tt]{gebauer2018generating}
N.~W.~A. Gebauer, M.~Gastegger, and K.~T. Sch{\"u}tt.
\newblock Generating equilibrium molecules with deep neural networks.
\newblock \emph{arXiv preprint arXiv:1810.11347}, 2018.

\bibitem[van~den Oord et~al.(2016)van~den Oord, Kalchbrenner, Espeholt,
  Kavukcuoglu, Vinyals, and Graves]{van2016conditional}
A.~van~den Oord, N.~Kalchbrenner, L.~Espeholt, K.~Kavukcuoglu, O.~Vinyals, and
  A.~Graves.
\newblock Conditional image generation with {PixelCNN} decoders.
\newblock In D.~D. Lee, M.~Sugiyama, U.~V. Luxburg, I.~Guyon, and R.~Garnett,
  editors, \emph{Advances in Neural Information Processing Systems 29}, pages
  4790--4798. Curran Associates, Inc., 2016.

\bibitem[Ramachandran and Varoquaux(2011)]{ramachandran2011mayavi}
P.~Ramachandran and G.~Varoquaux.
\newblock {Mayavi: 3D Visualization of Scientific Data}.
\newblock \emph{Computing in Science \& Engineering}, 13\penalty0 (2):\penalty0
  40--51, 2011.
\newblock ISSN 1521-9615.

\bibitem[Kingma and Ba(2014)]{kingma2014adam}
D.~P. Kingma and J.~Ba.
\newblock Adam: A method for stochastic optimization.
\newblock \emph{arXiv preprint arXiv:1412.6980}, 2014.

\bibitem[Schütt et~al.(2019)Schütt, Kessel, Gastegger, Nicoli, Tkatchenko,
  and Müller]{schutt2019schnetpack}
K.~T. Schütt, P.~Kessel, M.~Gastegger, K.~A. Nicoli, A.~Tkatchenko, and K.-R.
  Müller.
\newblock \mbox{SchNetPack}: A deep learning toolbox for atomistic systems.
\newblock \emph{Journal of Chemical Theory and Computation}, 15\penalty0
  (1):\penalty0 448--455, 2019.
\newblock \doi{10.1021/acs.jctc.8b00908}.

\bibitem[Paszke et~al.(2019)Paszke, Gross, Massa, Lerer, Bradbury, Chanan,
  Killeen, Lin, Gimelshein, Antiga, Desmaison, Kopf, Yang, DeVito, Raison,
  Tejani, Chilamkurthy, Steiner, Fang, Bai, and Chintala]{pytorch2019neurips}
A.~Paszke, S.~Gross, F.~Massa, A.~Lerer, J.~Bradbury, G.~Chanan, T.~Killeen,
  Z.~Lin, N.~Gimelshein, L.~Antiga, A.~Desmaison, A.~Kopf, E.~Yang, Z.~DeVito,
  M.~Raison, A.~Tejani, S.~Chilamkurthy, B.~Steiner, L.~Fang, J.~Bai, and
  S.~Chintala.
\newblock {PyTorch}: An imperative style, high-performance deep learning
  library.
\newblock In H.~Wallach, H.~Larochelle, A.~Beygelzimer, F.~d\textquotesingle
  Alch\'{e}-Buc, E.~Fox, and R.~Garnett, editors, \emph{Advances in Neural
  Information Processing Systems 32}, pages 8024--8035. Curran Associates,
  Inc., 2019.

\bibitem[Larsen et~al.(2017)Larsen, Mortensen, Blomqvist, Castelli,
  Christensen, Du{\l}ak, Friis, Groves, Hammer, Hargus, et~al.]{larsen2017ase}
A.~H. Larsen, J.~J. Mortensen, J.~Blomqvist, I.~E. Castelli, R.~Christensen,
  M.~Du{\l}ak, J.~Friis, M.~N. Groves, B.~Hammer, C.~Hargus, et~al.
\newblock The atomic simulation environment—a python library for working with
  atoms.
\newblock \emph{Journal of Physics: Condensed Matter}, 29\penalty0
  (27):\penalty0 273002, 2017.

\bibitem[O'Boyle et~al.(2011)O'Boyle, Banck, James, Morley, Vandermeersch, and
  Hutchison]{openbabel}
N.~M. O'Boyle, M.~Banck, C.~A. James, C.~Morley, T.~Vandermeersch, and G.~R.
  Hutchison.
\newblock Open babel: An open chemical toolbox.
\newblock \emph{Journal of Cheminformatics}, 3\penalty0 (1):\penalty0 33, Oct
  2011.

\bibitem[Becke(1993)]{Becke1993JCP}
A.~D. Becke.
\newblock Density-functional thermochemistry. {III.} the role of exact
  exchange.
\newblock \emph{J. Chem. Phys.}, 98:\penalty0 5648--5652, 1993.

\bibitem[Lee et~al.(1988)Lee, Yang, and Parr]{Lee1988PRBb}
C.~Lee, W.~Yang, and R.~G. Parr.
\newblock {LYP} correlation: Development of the {Colle-Salvetti}
  correlation-energy formula into a functional of the electron density.
\newblock \emph{Phys. Rev. B}, 37\penalty0 (2):\penalty0 785, 1988.

\bibitem[Vosko et~al.(1980)Vosko, Wilk, and Nusair]{Vosko1980CJP}
S.~H. Vosko, L.~Wilk, and M.~Nusair.
\newblock Accurate spin-dependent electron liquid correlation energies for
  local spin density calculations: a critical analysis.
\newblock \emph{Can. J. Phys.}, 58\penalty0 (8):\penalty0 1200--1211, 1980.

\bibitem[Stephens et~al.(1994)Stephens, Devlin, Chabalowski, and
  Frisch]{Stephens1994JPC}
P.~J. Stephens, F.~J. Devlin, C.~F. Chabalowski, and M.~J. Frisch.
\newblock Ab initio calculation of vibrational absorption and circular
  dichroism spectra using density functional force fields.
\newblock \emph{J. Phys. Chem.}, 98\penalty0 (45):\penalty0 11623--11627, 1994.

\bibitem[Neese(2012)]{Neese2012WCMS}
F.~Neese.
\newblock The {ORCA} program system.
\newblock \emph{WIREs Comput. Mol. Sci.}, 2\penalty0 (1):\penalty0 73--78,
  2012.

\bibitem[RDKit, online()]{rdkit}
RDKit, online.
\newblock {RDK}it: Open-source cheminformatics.
\newblock \url{http://www.rdkit.org}.
\newblock [Online; accessed 23-May-2019].

\bibitem[Perdew et~al.(1996)Perdew, Burke, and Ernzerhof]{Perdew1996PRL}
J.~P. Perdew, K.~Burke, and M.~Ernzerhof.
\newblock Generalized gradient approximation made simple.
\newblock \emph{Phys. Rev. Lett.}, 77:\penalty0 3865--3868, 1996.

\bibitem[Weigend and Ahlrichs(2005)]{Weigend2005PCCP}
F.~Weigend and R.~Ahlrichs.
\newblock Balanced basis sets of split valence, triple zeta valence and
  quadruple zeta valence quality for {H} to {Rn}: Design and assessment of
  accuracy.
\newblock \emph{Phys. Chem. Chem. Phys.}, 7\penalty0 (18):\penalty0 3297--3305,
  2005.

\bibitem[Eichkorn et~al.(1995)Eichkorn, Treutler, {\"O}hm, H{\"a}ser, and
  Ahlrichs]{Eichkorn1995CPL}
K.~Eichkorn, O.~Treutler, H.~{\"O}hm, M.~H{\"a}ser, and R.~Ahlrichs.
\newblock Auxiliary basis sets to approximate {C}oulomb potentials.
\newblock \emph{Chem. Phys. Lett.}, 240\penalty0 (4):\penalty0 283--290, 1995.

\bibitem[Vahtras et~al.(1993)Vahtras, Alml{\"o}f, and
  Feyereisen]{Vahtras1993CPL}
O.~Vahtras, J.~Alml{\"o}f, and M.~W. Feyereisen.
\newblock Integral approximations for {LCAO-SCF} calculations.
\newblock \emph{Chem. Phys. Lett.}, 213\penalty0 (5-6):\penalty0 514--518,
  1993.

\bibitem[Grimme et~al.(2010)Grimme, Antony, Ehrlich, and Krieg]{Grimme2010JCP}
S.~Grimme, J.~Antony, S.~Ehrlich, and H.~Krieg.
\newblock {A consistent and accurate \emph{ab initio} parametrization of
  density functional dispersion correction ({DFT-D)} for the 94 elements H-Pu}.
\newblock \emph{J. Chem. Phys.}, 132\penalty0 (15):\penalty0 154104, 2010.

\bibitem[Simonovsky and Komodakis(2018)]{simonovsky2018graphvae}
M.~Simonovsky and N.~Komodakis.
\newblock {GraphVAE}: Towards generation of small graphs using variational
  autoencoders.
\newblock In \emph{International Conference on Artificial Neural Networks},
  pages 412--422. Springer, 2018.

\bibitem[Samanta et~al.(2019)Samanta, Abir, Jana, Chattaraj, Ganguly, and
  Gomez-Rodriguez]{samanta2019nevae}
B.~Samanta, D.~Abir, G.~Jana, P.~K. Chattaraj, N.~Ganguly, and
  M.~Gomez-Rodriguez.
\newblock {NeVAE}: A deep generative model for molecular graphs.
\newblock In \emph{Proceedings of the AAAI Conference on Artificial
  Intelligence}, volume~33, pages 1110--1117, 2019.

\end{thebibliography}
	
	
	\newpage
	\section*{Supplementary material: Symmetry-adapted generation of 3d point sets for the targeted discovery of molecules}
	\label{sec:supplements}
	\addcontentsline{toc}{section}{Supplementary material}
	
	\subsection*{Architecture}
	\label{subsec:architecture}
	\addcontentsline{toc}{subsection}{Architecture}
	Here we summarize the exact settings used in all layers of our neural network architecture.
	The structure of the interaction blocks can be found in~\citet{schutt2017schnet}.
	The number of atom features was set to 128 and used in all atom-wise dense layers of the interaction block and filter-generating layers.
	Distances are expanded in the filter-generating layers using 25 Gaussians with equally spaced centers $0$~{\AA}$\leq\mu\leq10$~{\AA}.
	Overall, we use nine interaction blocks for feature extraction.
	We re-use one embedding layer with 128 features at all steps depicted in Fig.~\ref{fig:architecture} of the paper.
	The output network for type predictions consists of five atom-wise dense layers with shifted-softplus non-linearity and 128, 96, 64, 32, and 1 atom features, respectively.
	The output network for distance predictions also consists of five atom-wise dense layers with shifted-softplus non-linearity and 128, 171, 214, 257, and 300 atom features.
	Both output networks contain a final softmax layer.
	
	\subsection*{Training details}
	\label{subsec:training}
	\addcontentsline{toc}{subsection}{Training details}
	The neural networks were trained with stochastic gradient descent using the ADAM optimizer~\cite{kingma2014adam}.
	We start with a learning rate of $10^{-4}$ which is reduced using a decay factor of $0.5$ after $10$ epochs without improvement of the validation loss.
	The training is stopped at $\text{lr} \leq 10^{-6}$.
	Afterwards, the model with lowest validation error is selected for generation.
	
	While the atom type labels $\mathbf{q}_{i}^\text{type}$ can be directly obtained from the training data, the labels for distance distributions $\mathbf{q}_{ij}^\text{dist}$ are obtained using:
	\[
	\left[\mathbf{q}_{ij}^\text{dist}\right]_l = \frac{\exp(-\frac{1}{\gamma} (d_{(t+i)j} - \mu_l)^2)}{\sum_{l=1}^{300} \exp(-\frac{1}{\gamma} (d_{(t+i)j} - \mu_l)^2)} \hspace{0.6cm} \forall \hspace{0.1cm} j < t+i.
	\]
	The width of the Gaussians can be tuned with the $\gamma$ parameter, which we set to 10\% of the bin size in our experiments, resulting in very peaky, uni-modal label vectors.
	
	\subsection*{Controlling randomness with the temperature parameter}
	\label{subsec:randomness}
	\addcontentsline{toc}{subsection}{Controlling randomness with the temperature parameter}
	In order to control the randomness during generation, we do not directly implement Eq.~\ref{eq:dist_to_pos} but include a temperature parameter $T$:
	\begin{align}
	p(\mathbf{r}_{t+i}|\mathbf{R}^{t}_{\leq{i-1}},\mathbf{Z}^{t}_{\leq{i}}) = \frac{1}{\alpha} \exp\left(\frac{\sum_{j=1}^{t+i-1}\log p(d_{(t+i)j}|\mathbf{x}_j)}{T}\right).\label{eq:dist_pos_T}
	\end{align}
	Increasing $T$ will smoothen the grid distribution, effectively increasing randomness, whereas small values lead to a peaky distribution and thus less randomness.
	For all experiments, we chose a fixed temperature of $T=0.1$ according to the following procedure.
	
	We used the G-SchNet model trained on 50k equilibrium structures from QM9 \cite{ramakrishnan2014quantum,reymond2015chemical,ruddigkeit2012enumeration} and generated 20k molecules for each $T\in\{2, 1, 0.1, 0.01, 0.001\}$. From each set, 1k valid and unique molecules were randomly chosen, where 800 resembled test structures, 100 resembled training structures, and the remaining 100 were novel structures with more than 9 heavy atoms. We relaxed the five sets of 1k molecules at the PBE/def2-SVP level of theory\cite{Perdew1996PRL, Weigend2005PCCP} using the Orca program, employing the resolution of identity (RI) approximation\cite{Eichkorn1995CPL,Vahtras1993CPL} and Grimme D3 dispersion correction with Becke-Johnson damping.\cite{Grimme2010JCP} The root-mean-square deviation (RMSD) between atomic positions before and after relaxation was measured. A smaller RMSD means that the generated structures are closer to a true equilibrium configuration.
	
	\begin{figure}
		\centering
		\includegraphics[width=0.4\linewidth]{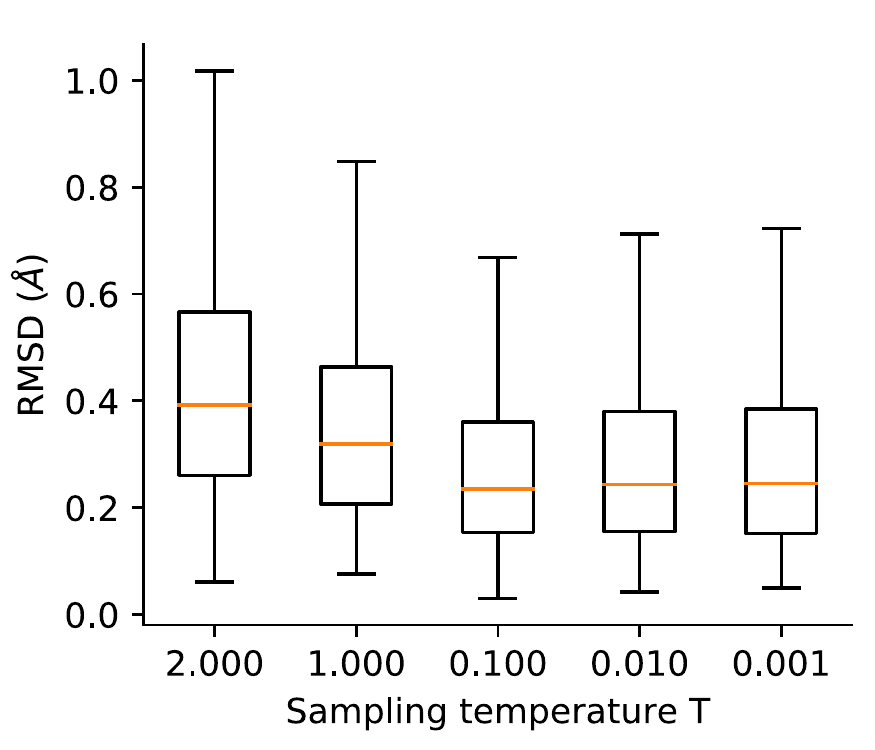}
		\caption{The boxplots show the RMSD of atomic positions between generated and relaxed structures for different values of $T$, where the boxes extend from the lower to the upper quartile values, the red lines indicate the medians, and the whiskers reach up to 1.5 times the interquartile ranges. Outliers are not shown.\label{fig:rmsd_t}}
	\end{figure}
	In Fig.~\ref{fig:rmsd_t} we show boxplots of the RMSD for the five different temperatures $T$. We see that the smallest median values and interquartile ranges are observed for values of $T$ smaller than $1$. Since increasing $T$ corresponds to increasing randomness during sampling, the increase in the RMSD for $T=1$ and $T=2$ is expected. However, decreasing $T$ to smaller values than $0.1$ does not lead to smaller RMSDs. Generally, we expect the number of unique sampless to decrease as $T$ gets too small. Therefore, for the other experiments, we chose the highest value for $T$ that still produces structures close to equilibrium, namely $T=0.1$.
	
	\subsection*{Sampling generation traces for training}
	\label{subsec:traces}
	\addcontentsline{toc}{subsection}{Sampling generation traces for training}
	The generation traces start with the focus and origin tokens set to the same position, which is the center of mass of the respective training molecule.
	The first new type and position are taken from the atom closest to the center of mass.
	At each subsequent step, we randomly select one of the already placed non-token atoms as focus point (which is only a single choice for the second step).
	The new position and type are then taken from the neighbor which is closest to the origin token, where neighbors are all unplaced atoms of the training molecule that are connected to the focus by a bond.
	If no unplaced neighbors are left, the new type is set to the stop token and the focused atom is marked as finished, i.e. it cannot be chosen as focus anymore.
	For the next step, another already placed (unfinished) atom is randomly chosen as focus and the procedure is repeated until all atoms have been placed and marked as finished.
	
	\subsection*{Generating molecules}
	\label{subsec:generation}
	\addcontentsline{toc}{subsection}{Generating molecules}
	For the first step of molecule generation, the focus and origin token are set to the origin of a 3d grid.
	The grid extends up to 1.7~{\AA} into all dimensions with a step-size of 0.05~{\AA}. 
	The token are processed by G-SchNet to sample the type of the first non-token atom and to obtain the two predicted distance distributions.
	The probabilities of the grid cell positions are calculated according to Eq.~\ref{eq:dist_pos_T} in order to sample the position of the first atom.
	Then, at each subsequent step, a random unfinished non-token atom is selected as focus token (which is only a single choice for the second step).
	The grid is centered on the focus (but the origin token stays at the former origin of the grid) and G-SchNet predictions are obtained to sample the type and position of the next atom (see Fig.~\ref{fig:architecture} in the paper for an exemplary generation step).
	If the stop token is predicted as next type, the currently focused atom is marked as finished and no position is sampled.
	Instead the next generation step is initialized by randomly selecting one of the remaining unfinished atoms as focus point.
	The generation process stops if no unfinished atoms are left.
	For our experiments, we also stopped the process if molecules were not finished after placing 35 atoms and discarded these structures as invalid (usually \textasciitilde1\% of generated molecules).
	
	\subsection*{Matching molecules}
	\label{subsec:matching}
	\addcontentsline{toc}{subsection}{Matching molecules}
	To remove non-unique structures and identify moleclues that resemble training or test data, we calculate the Tanimoto similarity of path-based fingerprints (FP2)~\cite{openbabel} for pairs of molecules with Open Babel.
	If the similarity is one, we compare the canonical SMILES representations in a second step.
	If they match, the two molecules are treated as equal.
	Note that this is a conservative approach as it filters out some spatial isomers which cannot be distinguished with SMILES.
	
	\subsection*{Ablation study}
	\label{subsec:ablation}
\begin{figure}
		\centering
		\includegraphics[width=0.95\linewidth]{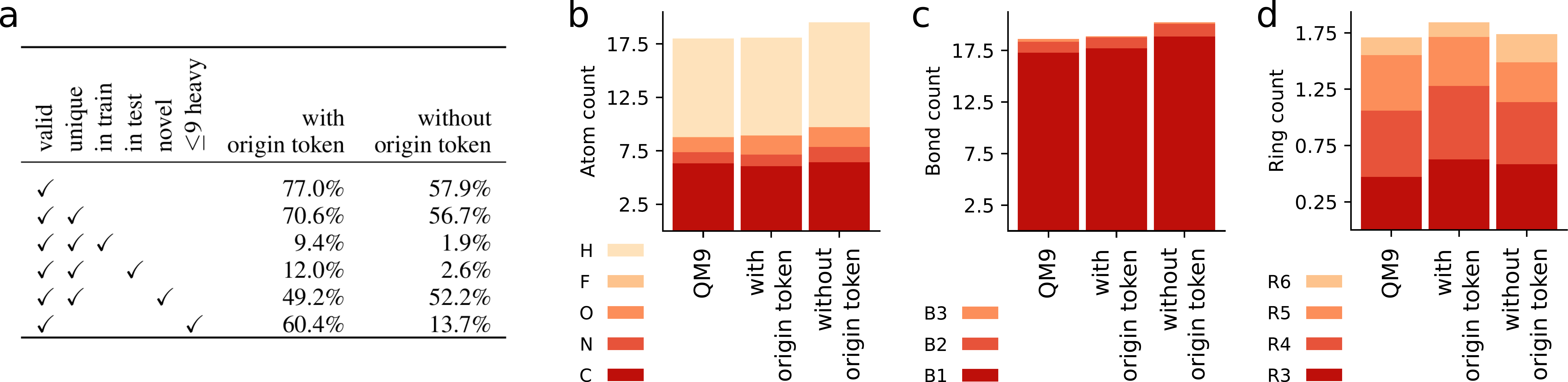}
		\caption{Ablation study on the effect of the origin token. Statistics are compared for 20k molecules generated by a G-SchNet model with origin token and 20k molecules generated by a G-SchNet model without origin token. Table~(a) shows the percentage of generated molecules for which the properties indicated by the check marks in each row hold. The average numbers of atoms, bonds, and rings per generated molecule and per QM9 molecule are compared in (b), (c), and (d), respectively. B1, B2, and B3 correspond to single, double, and triple bonds. R3, R4, R5, and R6 are rings of size 3 to 6.\label{fig:ablation}}
	\end{figure}

	\addcontentsline{toc}{subsection}{Ablation study}
	In order to assess the effect of the origin token, we conduct an ablation study where we remove the origin token from the input to a G-SchNet model during training and generation.
	All other hyperparameters are identical to the ones used when training the standard G-SchNet with origin token.
	After training, we generate 20k molecules with each architecture and compare their statistics.
	Fig.~\ref{fig:ablation}a shows that the validity of generated molecules drops by almost 20 percent without the origin token.
	Furthermore, the amount of generated molecules that match QM9 training or test structures significantly decreases.
	All QM9 structures consist of at most 9 heavy atoms but only 13.7\% of the molecules generated without origin token have 9 or less heavy atoms (compared to 60.4\% with origin token).
	The diverging atom count can also be seen in Fig.~\ref{fig:ablation}b.
	Moreover, the bond count in Fig.~\ref{fig:ablation}c also diverges from the training data distribution.
	The ring count in Fig.~\ref{fig:ablation}d, on the other hand, is not noticeably better or worse without origin token.
	There is an increase in six-membered rings and a decrease in all smaller ring structures compared to the model with origin token.
	Both models slightly diverge from the QM9 training data ring count.
	Overall, we conclude that the origin token has a significant, positive effect on the approximated probability distribution.
	It enables G-SchNet to better capture the characteristics of the training data, leading to a model that generates more valid molecules which are more faithful to the training structures but still equally unique and unseen (in test or novel) compared to a model without origin token. 
	
	\subsection*{Targeted discovery with respect to further electronic properties}
	\label{subsec:discovery}
\begin{figure}
		\centering
		\includegraphics[width=0.95\linewidth]{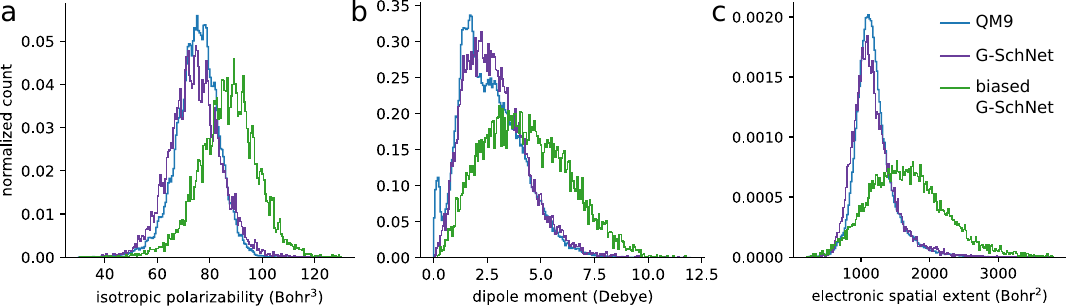}
		\caption{Distribution of three quantum-chemical properties for molecules from QM9 (blue), generated by an unbiased G-SchNet (purple), and generated by G-SchNets biased towards larger values of the respective property (green).\label{fig:bias}}
	\end{figure}

	\addcontentsline{toc}{subsection}{Targeted discovery with respect to further electronic properties}
	In a similar fashion to the experiments for molecules with a small HOMO-LUMO gap, we bias G-SchNet models towards large values of three more electronic properties available in QM9, namely the isotropic polarizability, the dipole moment, and the electronic spatial extent.
	The results depicted in Fig.~\ref{fig:bias} show clear shifts in the distribution of targeted properties for molecules generated with biased G-SchNets.
	We again use small subsets of molecules from QM9 exhibiting the respective property for fine-tuning of the G-SchNet model previously trained on 50k structures.
	These subsets consist of $2100/500$ molecules with an isotropic polarizability $\geq 91$ Bohr$^3$, $3000/500$ molecules with a dipole moment $\geq 5.75$ Debye, and $4400/500$ molecules with an electronic spatial extent $\geq 1785$ Bohr$^2$ for training/validation, respectively.
	In contrast to the HOMO-LUMO gap experiments, where we relaxed generated structures and calculated the gap numerically with time-consuming DFT simulations, we train three separate SchNet models to predict the three electronic properties.
	We use 100k molecules from QM9 for training, 10k for validation, and the remaining structures as a test set.
	The mean absolute test error is $0.070$ Bohr$^3$ for the electronic polarizability, $0.016$ Debye for the dipole moment, and $0.126$ Bohr$^2$ for the electronic spatial extent.
	
	\subsection*{Detailed statistics}
	\label{subsec:statistics}
	\addcontentsline{toc}{subsection}{Detailed statistics}
	We provide two tables which report more detailed statistics on relevant properties of generated molecules.
	In Table~\ref{tbl:stats} we compare 20k molecules generated by our standard G-SchNet model, by the G-SchNet model trained on QM9 without structures with 3- or 4-membered rings, by the G-SchNet model biased towards small HOMO-LUMO gaps, and by the constrained graph variational autoencoder (CGVAE).
	The provided numbers are percent of all validly generated structures.
	
	In Table~\ref{tbl:comparison} we compare the number of valid, novel, and unique molecules generated by the standard G-SchNet as well as six related generative models that rely either on graphs or SMILES strings as molecule representation.
	The shown numbers are the respective percentage of 20k generated molecules for each model.
	Note that this means that a high novelty score can also come from a low number of valid molecules, as each invalid structure is not included in the training data and therefore novel.
	Similarly, a high uniqueness score is only interesting if the majority of generated molecules is valid as otherwise the model generates many unique but invalid structures.
	In general, the methods cannot directly be compared since generating 3d molecular structures is a different task than generating graphs or strings.
	For example, all of the valid G-SchNet molecules correspond to one proper 3d structure whereas valid graphs and SMILES strings may have no, one, or many different corresponding 3d conformers which cannot easily be found without expensive quantum-chemical simulations.
	
	\begin{table}
		\caption{Statistics for all of our models and CGVAE~\cite{liu2018constrained} (molecules provided by the authors). CGVAE was trained on all molecules in QM9. Our models were trained on 50k randomly selected molecules from QM9. For the second model molecules with three- and four-membered rings were excluded. The third model was fine-tuned on 3k molecules with a HOMO-LUMO gap $\leq$ 4.5~eV.  The numbers are percent of all validly generated molecules (which are 77\%, 78\%, 69\% and 100\% of 20k generated for the models from left to right). The molecules are categorized according to the checkmarks in the first five columns. Unseen refers to molecules not found in the training data and novel stands for molecules not found in QM9. $\leq$9 heavy marks molecules with nine or less heavy atoms and gap refers to the HOMO-LUMO gap and was calculated for relaxed structures.}
		\label{tbl:stats}
		\centering
		\renewcommand{\arraystretch}{0.01}
		\begin{tabular}{C{0.01cm}C{0.01cm}C{0.01cm}C{0.01cm}C{0.01cm}R{2cm}R{2.3cm}R{2.5cm}R{1cm}|R{1.5cm}}
			\toprule
			\rotatebox{90}{unique} & 
			\rotatebox{90}{unseen} &
			\rotatebox{90}{novel} &
			\rotatebox{90}{$\leq$9 heavy}&
			\rotatebox{90}{gap $\leq$5eV} & 
			G-SchNet\newline50k &
			G-SchNet\newline50k no R3/R4&
			G-SchNet\newline50k +\newline3k gap$\leq$4.5eV & &
			CGVAE\newline\textasciitilde134k\\
			\midrule
			\checkmark & && & & 91.6\% & 89.4\% & 73.8\% && 98.4\%\\
			\checkmark & \checkmark && & & 79.4\% & 68.3\% & 66.8\% && 87.4\%\\
			\checkmark & & \checkmark & & & 63.8\% & 68.3\% & 57.5\% && 87.4\%\\
			\checkmark &  && \checkmark &  & 70.2\% & 58.2\% & 57.8\% && 30.0\%\\
			\checkmark & \checkmark && \checkmark &  & 58.0\% & 37.1\% & 50.8\% && 18.9\%\\
			\checkmark && \checkmark & \checkmark &  & 42.4\% & 37.1\% & 41.5\% && 18.9\%\\
			\checkmark & \checkmark && & \checkmark & 6.8\% & 21.2\%& 43.4\% && --\\
			\bottomrule
		\end{tabular}
		\renewcommand{\arraystretch}{1}
	\end{table}
	
	\begin{table}[h!]
		\begin{center}
			\caption{Percent of valid, novel (not in training data), and unique molecules among 20k structures generated after training on QM9 for G-SchNet and related models working with SMILES or graph representations. Molecules are considered valid if the valency-constraints of all its atoms are met. In order to identify duplicate and novel molecules, we use molecular fingerprints and canonical SMILES strings as explained in section "matching molecules" above. Statistics for graph- and SMILES-based models are taken from Fig. 3 in \citet{liu2018constrained}.}
			\label{tbl:comparison}
			\begin{adjustbox}{max width=\textwidth}
				\begin{tabular}{l|r|r|r|r|r|r|r}
					\toprule 
					\textbf{}& 
					\textbf{G-SchNet}& 
					\textbf{CGVAE}*\cite{liu2018constrained}& 
					\textbf{GraphVAE}*\cite{simonovsky2018graphvae}& 
					\textbf{NeVAE}*\cite{samanta2019nevae}&
					\textbf{LSTM}*\cite{liu2018constrained}& 
					\textbf{CVAE}*\cite{gomez2016automatic}& 
					\textbf{GVAE}*\cite{kusner2017grammar}\\
					& 3d structure & graph & graph & graph & SMILES & SMILES & SMILES \\
					\midrule
					\textbf{valid} & 77.07\% & 100.00\% & 61.00\% & 98.00\% & 94.78\% & 10.00\% & 30.00\%\\
					\textbf{novel} & 87.47\% & 94.30\% & 85.00\% & 100.00\% & 82.98\% & 90.00\% & 95.44\%\\
					\textbf{unique} & 91.91\% & 98.57\% & 40.90\% & 99.86\% & 96.94\% & 64.50\% & 9.30\%\\
					\bottomrule 
					\multicolumn{8}{r}{*statistics are taken from Fig. 3 in \citet{liu2018constrained}}
				\end{tabular}
			\end{adjustbox}
		\end{center}
	\end{table}
	
	
\end{document}